\documentclass{article}

     \PassOptionsToPackage{numbers, compress}{natbib}
\usepackage[preprint]{neurips_2024}

\usepackage[utf8]{inputenc} % allow utf-8 input
\usepackage[T1]{fontenc}    % use 8-bit T1 fonts
\usepackage{hyperref}       % hyperlinks
\usepackage{url}            % simple URL typesetting
\usepackage{booktabs}       % professional-quality tables
\usepackage{amsfonts}       % blackboard math symbols
\usepackage{nicefrac}       % compact symbols for 1/2, etc.
\usepackage{microtype}      % microtypography
\usepackage{xcolor}         % colors
\usepackage{graphicx}
\usepackage{amsmath}

\usepackage{multirow}

\usepackage{listings}

\title{\textit{This} Looks Better than \textit{That}: Better Interpretable Models with ProtoPNeXt}

\author{%
  Frank Willard$^{*}$ \\
  Department of Computer Science\\
  Duke University\\
  Durham, NC \\
  \texttt{Frank.Willard@duke.edu} \\
  \And
  Luke Moffett$^{*c}$ \\
  Department of Computer Science\\
  Duke University\\
  Durham, NC \\\texttt{Luke.Moffett@duke.edu} \\
  \And
  Emmanuel Mokel \\
  Department of Computer Science\\
  Duke University\\
  Durham, NC \\\texttt{Emmanuel.Mokel@duke.edu} \\
  \And
  Jon Donnelly \\
  Department of Computer Science\\
  Duke University\\
  Durham, NC \\
  \texttt{Jon.Donelly@duke.edu} \\
  \And
  Stark Guo \\
  Department of Computer Science\\
  Duke University\\
  Durham, NC \\\texttt{Stark.Guo@duke.edu} \\
  \And
  Julia Yang \\
  Department of Computer Science\\
  Duke University\\
  Durham, NC \\\texttt{Julia.Yang@duke.edu} \\
  \And
  Giyoung Kim \\
  Department of Computer Science\\
  Duke University\\
  Durham, NC \\\texttt{Giyoung.Kim@duke.edu} \\
  \And
  Alina Jade Barnett \\
  Department of Computer Science\\
  Duke University\\
  Durham, NC \\\texttt{Alina.Barnett@duke.edu} \\
  \And
  Cynthia Rudin \\
  Department of Computer Science\\
  Duke University\\
  Durham, NC \\\texttt{Cynthia.Rudin@duke.edu}
}

\begin{document}

\maketitle
\let\thefootnote\relax\footnotetext{$^{*}$Co-First Authors, $^{c}$Corresponding Author}
% \blfootnote{$^{*}$Co-First Authors, $^{c}$Corresponding Author}

\begin{abstract}
Prototypical-part models are a popular interpretable alternative to black-box deep learning models for computer vision. 
However, they are difficult to train, with high sensitivity to hyperparameter tuning, inhibiting their application to new datasets and our understanding of which methods truly improve their performance.
To facilitate the careful study of prototypical-part networks (ProtoPNets), we create a new framework for integrating components of prototypical-part models -- ProtoPNeXt.
Using ProtoPNeXt, we show that applying Bayesian hyperparameter tuning and an angular prototype similarity metric to the original ProtoPNet is sufficient to produce new state-of-the-art accuracy for prototypical-part models on CUB-200 across multiple backbones.
We further deploy this framework to jointly optimize for accuracy and prototype interpretability as measured by metrics included in ProtoPNeXt. 
Using the same resources, this produces models with substantially superior semantics and changes in accuracy between +1.3\% and -1.5\%. The code and trained models will be made publicly available upon publication.
\end{abstract}

\section{Introduction}
\label{sec:introduction}
Deep neural networks (DNN's) have become the dominant model class for a wide range of applications, particularly in computer vision. These models are generally accurate, yet uninterpretable, making them unsuitable for application in high stakes domains \cite{rudin2019stop}.

Recently, a line of work aiming to address this shortcoming through \textit{case-based} DNN's has emerged. This work began with ProtoPNet \cite{chen2019looks}, but has since produced myriad extensions of the original model (e.g., \cite{wang2021interpretable, donnelly2022deformable, nauta2021neural, nauta2023pip, rymarczyk2020protopshare, rymarczyk2022interpretable, barnett2021case, ma2024looks, sacha2023protoseg, kenny2022towards, wang2023learning}), and has been applied to domains as disparate as Alzheimer's disease \citep{wolf2023don}, breast cancer \citep{barnett2021case}, and deep fake detection \cite{bouter2023protoexplorer}. A number of these works \cite{wang2021interpretable, donnelly2022deformable, nauta2021neural} have demonstrated accuracy on par with or superior to black-box models, providing the interpretability that black-box models lack without sacrificing performance. They also showed through empirical studies that humans found these models understandable and easier to troubleshoot than black boxes. 
% The transition below is great if we want to focus on code...
Nonetheless, machine learning practitioners tend to default to using traditional, uninterpretable DNN's when approaching a new problem. If case-based models are effective and interpretable, why are they not used?

We propose that a significant reason is that the literature does not understand which factors are key to performance in these models, making effective deployment difficult.
%We propose that the adoption of case-based models has been hindered by the extensive tuning required to apply such models to novel datasets.
% We propose that the literature has not fully elucidated which changes have led to recent improvements in effectiveness of these models. 
% Because these models rely on discrete updates in training, they can be sensitive to hyperparameters, and 
% driven by a similar lack of comprehensive evaluation in black boxes.
The lack of systematic hyperparameter tuning that has become standard in the case-based DNN literature makes it unclear whether recent gains in performance are really due to improved methods, or are simply an outcome of better tuning.
This is particularly concerning because model configurations for case-based DNN's can affect not just performance but also model interpretability.
As a result, the research community may be overestimating the difficulty of using case-based DNN's and underestimating their potential.

\begin{figure}
    \centering
    \includegraphics[width=0.95\textwidth]{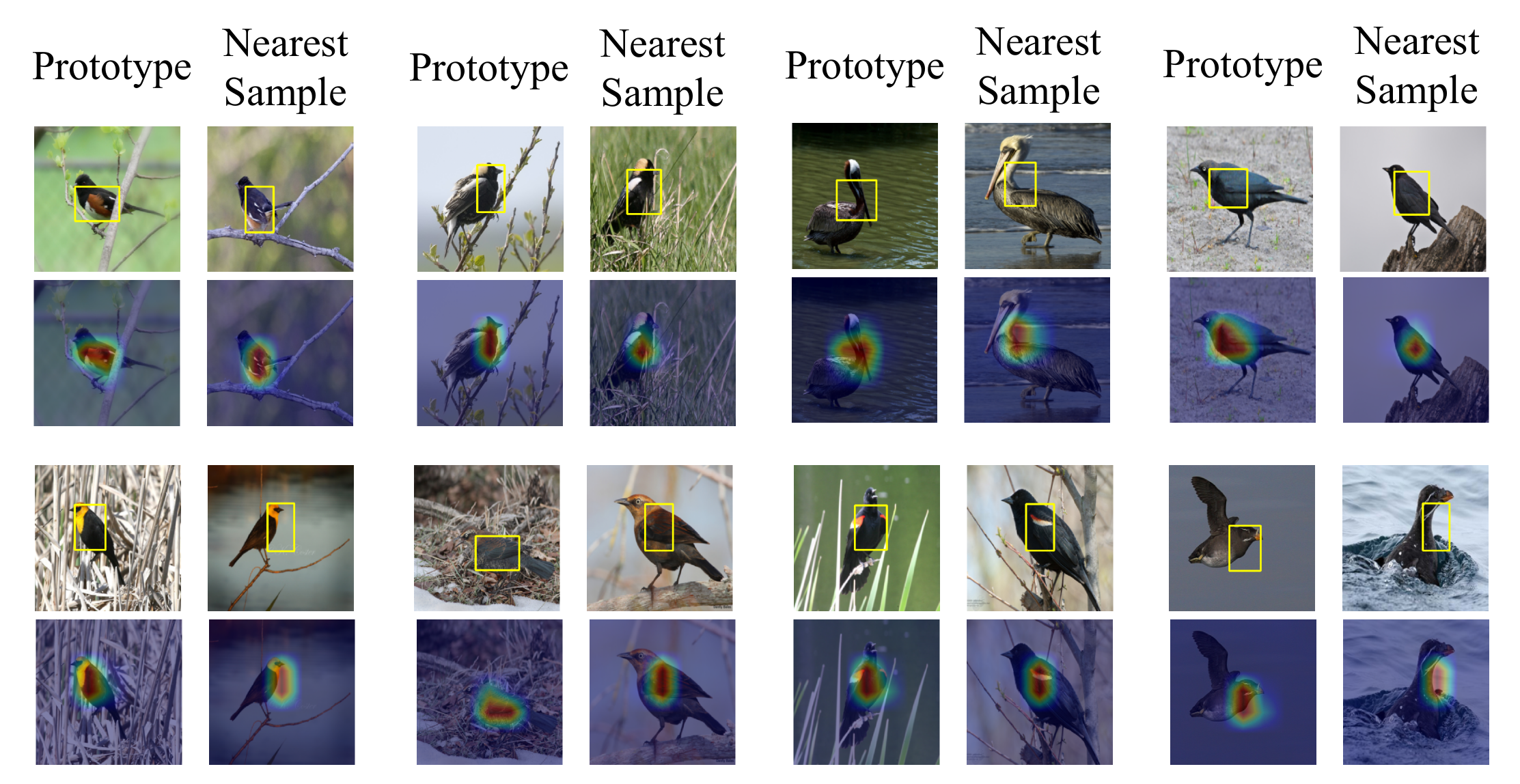}
    \caption{Eight randomly selected prototypes from a model produced with ProtoPNeXt. All prototypes from the model demonstrate strong semantics. The model has 86.2\% accuracy on uncropped CUB-200, and the full set of 253 prototypes from this model can be seen here: \url{https://drive.google.com/drive/folders/174f2PPhLRarLevOhjm3Vc_YDt8x11mQ-}.}
    \label{fig:proto-reasoning}
\end{figure}

In this work, we aim to better understand what leads to performant case-based DNN's while making them accessible to the broader machine learning community. In pursuit of this goal, we provide the following contributions:
\begin{enumerate}
    \item We investigate whether recent gains in accuracy among case-based DNN's are due to new methods, or simply better hyperparameter tuning. To do so, we introduce a unified framework for conceptualizing and implementing the disparate descendants of ProtoPNet, which we use to implement multiple ProtoPNet variations.
    \item We introduce a novel sparsity metric and stopping criterion for case-based DNN's, which allows us to perform fixed-GPU-computation experiments on these ProtoPNet variations via Bayesian hyperparameter optimization. 
    \item We demonstrate that changing the \textit{original} ProtoPNet to use an angular similarity measure and systematically tuning hyperparameters is sufficient to deliver state-of-the-art accuracy on CUB-200, suggesting that recent improvements in performance among case-based DNN's may be misattributed. 
    \item We define a joint accuracy and prototype quality optimization objective. Again applying Bayesian optimization under fixed computational cost, we show quantitative and qualitative improvements to interpretability can also be accomplished through systematic optimization with minimal cost to accuracy.
\end{enumerate}

\section{Related Work}
\label{sec:related_work}
Concerns over the black-box nature of DNN's have inspired two substantial bodies of work. The first studies \textit{post-hoc} explanation methods, which try to explain the reasoning of a black-box model. Saliency maps \cite{bach2015pixel,simonyan2013deep,sundararajan2017axiomatic} use the gradient of an output with respect to input pixels to highlight the most ``salient'' input features. Activation maximization methods \cite{erhan2009visualizing,nguyen2016synthesizing} study models by synthesizing input images that maximize the activation of a target neuron. Additionally, general purpose variable importance methods like SHAP \cite{lundberg2017unified} and permutation importance metrics inspired by \cite{breiman2001random} have been applied to computer vision. Concept activation methods \cite{kim2018interpretability} use auxiliary ``concept'' datasets to determine which high-level concepts are important for classification. For a more thorough review of \textit{post-hoc} explanation methods, see \cite{minh2022explainable, linardatos2020explainable}. Post-hoc methods have difficulties that are hard to overcome, including unfaithful and incomplete explanations \citep{rudin2019stop}.

Rather than attempting to explain black-box models, the second family of work aims to develop inherently interpretable alternative models with comparable performance to black-boxes. Our work focuses on \textit{case-based} DNN's, as introduced by \cite{li2018deep,chen2019looks}. ProtoPNet was introduced by Chen et al. \cite{chen2019looks}, which learns a set of prototypical parts for each class, and forms predictions for each new instance by comparing each learned prototype to the image. This reasoning process has been described as a ``this looks like that'' process. ProtoPNet is appealing because its explanations are faithful to the underlying predictions and it provides much richer information than simple saliency -- each part of its reasoning process can be audited as a visible comparison followed by a simple product and sum of numbers.

Since the release of ProtoPNet, a disparate array of extensions of the original method have been developed. Several models alter the mechanism by which prototype activations are translated into class predictions: ProtoTree \cite{nauta2021neural} forms predictions by using the similarity of its prototypes to traverse a soft decision tree, and ProtoPShare \cite{rymarczyk2020protopshare} and ProtoPool \cite{rymarczyk2022interpretable} introduce a prototype sharing mechanism that allows a given prototype to be shared by multiple classes. In another direction, works like TesNet \cite{wang2021interpretable}, which encourages prototypes to form orthogonal bases for class subspaces; PIP-Net \cite{nauta2023pip}, which introduced a self-supervised loss; and IAIA-BL \cite{barnett2021case}, which introduced a fine annotation loss to guide prototype activations,  innovate by introducing better loss terms for the optimization of these networks. Deformable ProtoPNet \cite{donnelly2022deformable} extended the prototypes of ProtoPNet to allow spatial deformations. ProtoConcept \cite{ma2024looks} alters how prototypes are interpreted, visualizing ``prototypical concepts'' using all image patches within some radius of a prototype rather than using only the nearest. Each of these papers have demonstrated improvements to some combination of predictive performance and model interpretability. Thus, we have the question we investigate here: what was necessary to achieve the recent improvements in ProtoPNets?

\section{Methods}
\label{sec:methods}
Let $\mathbf{X} \in \mathbb{R}^{C \times H \times W}$ denote an input image of height $H$ and width $W$ with $C$ channels, and let $\mathbf{Y} \in \{0, ..., K-1\}$ denote the class label of the image. A ProtoPNet consists of three primary components: an embedding layer $f: \mathbb{R}^{C \times H \times W} \to \mathbb{R}^{D \times H' \times W'},$ which extracts a $H'$ by $W'$ map of $D$ dimensional feature vectors from an image; a prototype layer $g: \mathbb{R}^{D \times H' \times W'} \to \mathbb{R}^{P \times H'' \times W''},$ which computes the similarity of each of $P$ learned prototypes to the input at each of $H''$ by $W''$ center locations; and a class prediction head $h: \mathbb{R}^{P \times H'' \times W''} \to \mathbb{R}^K,$ which uses these prototype similarities to compute an output logit for each class. While the specific form of each of these components changes, all existing extensions of ProtoPNet fall into this framework.

These models minimize a loss function of the form
\begin{align}
    \ell_{\text{overall}} (f, g, h, \mathbf{X}, \mathbf{Y}) = CE(h \circ g \circ f(\mathbf{X}), \mathbf{Y}) + \ell_{\text{interp}}(f, g, h, \mathbf{X}, \mathbf{Y}, h \circ g \circ f(\mathbf{X})),
\end{align}
where the overall loss $\ell_{\text{overall}}$ is the sum of the standard cross entropy loss, denoted $CE$, and a weighted sum over other loss terms aiming to encourage interpretability, denoted collectively as $\ell_{\text{interp}}.$ The terms in $\ell_{\text{interp}}$ generally encourage the latent embedding space produced by $f$ to be well clustered by class, and encourage prototypes to recover representative samples from each class.

In almost all ProtoPNet variants, a prototype \texttt{projection} step is performed during and at the end of training to associate each learned prototype with a specific pixel space visualization\footnote{ProtoConcept \cite{ma2024looks} forgoes this step, instead using balls in the latent space produced by $f$ as prototypes. Any image patch within the radius of a given prototype ball is considered an equally good representation of that prototype, so each prototype is associated with a set of visualizations.}, an operation we implement within the prototype layer.

\subsection{Standardized Interface for Prototype Models}
\label{subsec:interface}
\begin{figure}[t]
    \centering
    \includegraphics[width=0.75\textwidth]{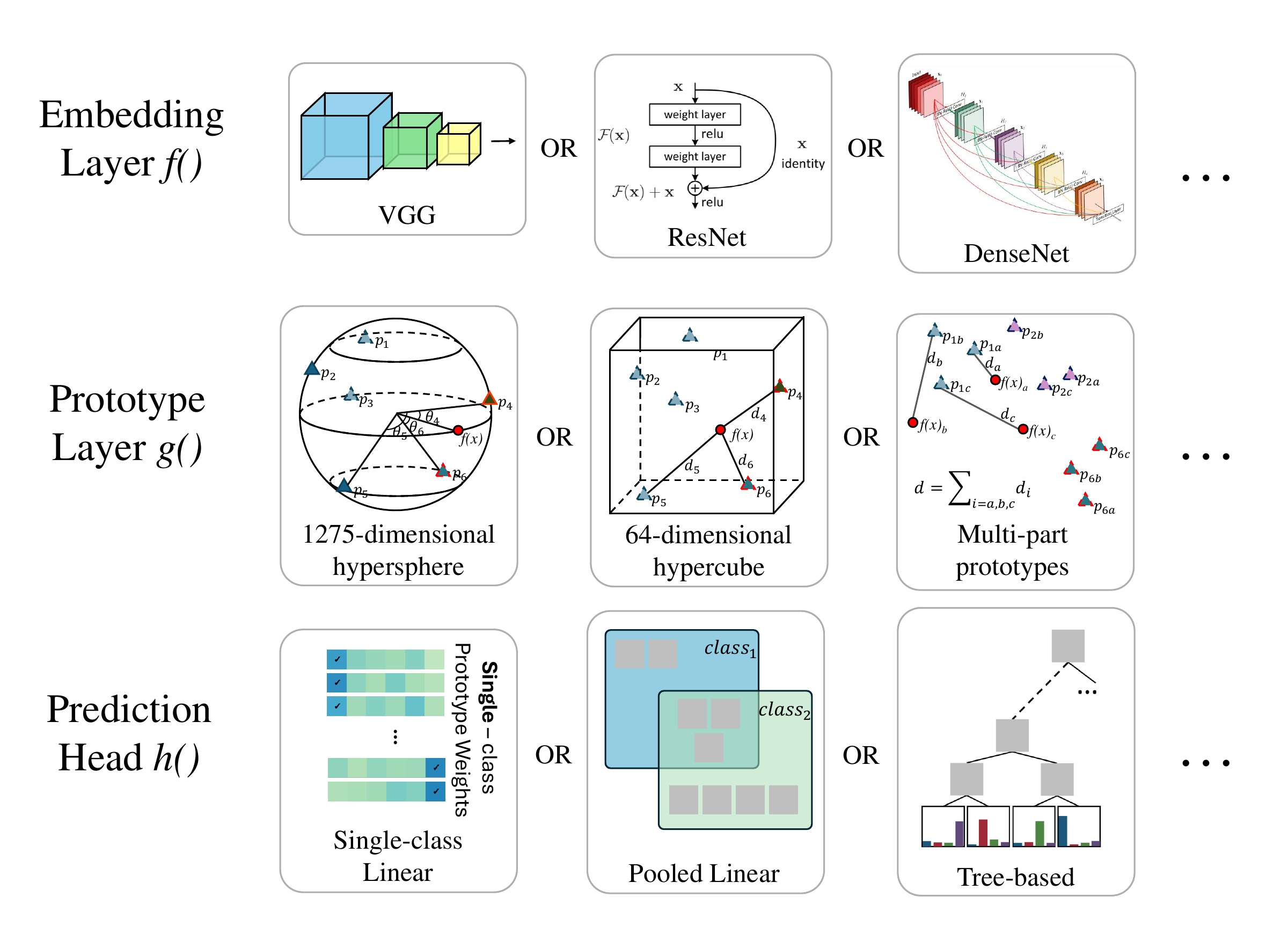}
    \caption{A high level overview of the ProtoPNext interface. A ProtoPNext is composed of three modules: an embedding layer, a prototype layer, and a prediction head. The embedding layer can be any black box vision model. The prototype layer compares learned, interpretable prototypes to the latent representation of the input. ProtoPNet \cite{chen2019looks} and Deformable ProtoPNet \cite{donnelly2022deformable} provide two variants of the prototype layer. The hypersphere visualization is adapted with permission from \cite{barnett2023interpretable}. The prediction head aggregates prototype activation values into a class prediction. ProtoPNet proposed a linear prediction head (bottom left), and alternative prediction heads have been proposed in ProtoTree \cite{nauta2021neural} (bottom right) and ProtoPool \cite{rymarczyk2022interpretable} (bottom middle).}
    \label{fig:protopnext-architecture}
\end{figure}

% Cynthia: ??? Add something about the question you're investigating, but also you want a useful modular interface.??
To determine which methodological changes actually improve case-based DNN performance, we need a simple, modular interface to support multiple extensions of ProtoPNet.
We propose such an interface, an overview of which is shown in Figure \ref{fig:protopnext-architecture}. A case-based DNN consists of three primary modules: an embedding layer, a prototype layer, and a prediction head. Embedding layers include a wide array of CNNs and vision transformers, optionally appended with some (possibly 0) number of additional convolutional layers. Existing prototype layers compare a set of learned parameters to the latent representation of an input under a specified metric, and include the original prototype from \cite{chen2019looks} and the deformable prototype from \cite{donnelly2022deformable}. Prediction heads include the linear layer from \cite{chen2019looks}, the tree from \cite{nauta2021neural}, and the pooling mechanism from \cite{rymarczyk2022interpretable}. 
%Thus, constructing a ProtoPNet that uses cosine similarity is a matter of changing the prototype layer.
In our implementation we separate the code into these conceptual units, reimplement ProtoPNet to use pure PyTorch (proprietary license) (including reimplementing deformable prototypes to avoid compiling custom CUDA-C++ code), and tune the training loop to avoid unnecessary computation. To support the complexity of this large, flexible code base, we implement an automated test suite that enables further expansion of the codebase features while ensuring core features are not broken. This framework can be readily extended and applied beyond our present investigation. 

Additionally, a fair investigation of these methods requires a systematic way to tune and select hyperparameters, which can be computationally expensive. In addition to the optimizations listed above, we address this by implementing a Bayesian optimization approach, described in detail in Appendix \ref{subsec:bayesian-optimization}. Beyond simply optimizing for accuracy, we are interested in maximizing the interpretability of these models wherever possible.
% Although Bayesian hyperparameter tuning can improve accuracy, a na\"ive implementation risks sacrificing interpretabilty if doing so would improve model accuracy. For example, for a dataset on which overfitting is unlikely, a ProtoPNet with $100$ prototypes per class may produce superior accuracy. However, this gain would come at the cost of a dense, difficult to understand model. 
As such, we implement and directly optimize several interpretability metrics, as described below.

\subsection{Prototype Metrics}
\label{subsec:prototype-metrics}
We compute three measures of model interpretability: a novel prototype sparsity metric, and stability and consistency, as introduced in \cite{huang2023evaluation}. In order to jointly optimize these metrics alongside accuracy, we ensure that each metric has a range of [0, 1], placing them on a scale consistent with that of accuracy.
All metrics are implemented as reusable TorchMetrics \cite{torchmetrics}.

\textbf{Prototype Sparsity.} Prototype Sparsity, introduced in this work, quantifies the amount of unique prototypes used by a model. To quantify sparsity, we need to account for the number of unique prototypes in a model, which determines how many images a user must consider when examining a given prediction, and the proportion of each image that the user must consider for each prototype.  

Recall that $K$ denotes the number of classes in the dataset, $H'$ the height of the latent space, and $W'$ the width of the latent space. For a model with prototype tensor $\mathbf{P} \in \mathbb{R}^{P \times D \times H_P \times W_P}$ (that is, the model has $P$ prototypes of spatial size $H_P \times W_P$), we compute a sparsity score $v_{\text{sparse}}$ as:
\begin{align}
    v_{\text{sparse}} = \left(K + \frac{K}{H'W'}\right) \big/ \left(P + \frac{P H_P W_P}{H' W'}\right).
\end{align}

This construction captures four desirable properties: 
First, sparsity and the number of prototype images are inversely proportional; doubling the number of prototype images halves the sparsity.
Second, a model with $K$ classes will have a sparsity of $1$ if it has $K$ prototype images and $K$ prototypical parts -- one prototype image per class and one prototypical part from that prototype image.
Third, adding a new prototypical part from an existing prototype image (which increases the term $H_PW_P/H'W'$) always decreases sparsity less than adding a new prototype image (which increases both terms in the denominator).
Its range is $[0,1]$, for easy comparison to other metrics.

\textbf{Prototype Stability and Consistency.} Prototype stability, $v_{\text{stab}}$, measures the invariance of prototype activation to Gaussian noise added to the input image, with perfectly stable prototypes demonstrating no change in activation in response to noise. Prototype consistency, $v_{\text{consist}}$, measures how frequently a prototype activates on the same semantic part of the image. For instance, a consistent prototype will always activate on the head of a bird, not the head and the feet.
Both metrics are formally defined in \cite{huang2023evaluation}.

\textbf{Optimization Objectives.} In separate experimental runs, we optimize for only accuracy (denoted Acc) $\text{obj}_{\text{acc}} = v_{\text{acc}}$, or for both accuracy and prototype quality scores (denoted Acc-PS). When directly optimizing interpretability,
we define a single prototype score as the average of the three prototype metrics,
\begin{align}
\label{eq:proto_score}
v_{\text{proto\_score}}
=\frac{(v_{\text{sparse}} + v_{\text{consist}} + v_{\text{stab}})}{3}.
\end{align}
When optimizing both accuracy and interpretability, we aim to maximize the following quantity:
\begin{align}
\label{eq:joint_obj}
    \text{obj}_\text{aps} = v_{\text{acc}} \cdot v_{\text{proto\_score}}.
\end{align}
This construction emphasizes \textit{complementarity} between accuracy and prototype quality in training prototypical-part models.

% Performance is assessed on two different measures. In \autoref{subsec:acc-experiments}, we optimize only for accuracy. In \autoref{subsec:interp-experiments}, we optimize jointly for accuracy and interpretability, as measured by the prototype score described in \autoref{subsec:prototype-metrics}.

\subsection{Prototype-Aware Early Stopping}
\label{subsec:early-stopping}
While training the embedding layers of a prototypical-part model, the prototypical parts, which exist in latent space, may decouple from their source images.
The projection step of training, in which each prototype is forced to be exactly equal to the most similar latent embedding from the training set, ties these prototypical parts to their source images.
During early phases of training, this projection may cause degradation in model performance that stabilizes over time.
This makes traditional early stopping, which relies on saturation of a target metric, unsuitable for prototypical-part models.

To address this, we introduce a novel early stopping scheme specific to training prototypical-part models.
We define \textit{patience} as the number of projection epochs without improvement.
We track a single metric, validation accuracy for epoch $e$, $v_{\text{acc}, e}$, at two separate points: (1) during projection epochs, $v_{\text{acc}}^{\text{proj}}$, and (2) in the epoch immediately before projection, $v_{\text{acc}}^{\text{preproj}}$.
Patience is exhausted only when the following condition is met during a specified number of sequential projection epochs, $p$:
\begin{align}
\label{eq:early_stop}
v_{\text{acc}, p} \leq \max(v_{\text{acc}}^{\text{proj}}) \text{ and } v_{\text{acc}, p-1} \leq \max(v_{\text{acc}}^{\text{preproj}}).
\end{align}
In our experiments, we use a projection patience of 3. We halt training once patience is exhausted.

\section{Evaluating What Causes Improvements in Accuracy}
\label{sec:experiments}
To study the source of accuracy improvements in recent ProtoPNet models, we first perform optimization with the accuracy-only objective.
We applied Bayesian hyperparameter tuning to a family of models with the features of both ProtoPNet \cite{chen2019looks} and Deformable ProtoPNet \cite{donnelly2022deformable} available, evaluating this family to more-challenging uncropped images from the CUB-200 image classification dataset \cite{WahCUB_200_2011}.
The CUB-200 dataset provides standardized train and test splits; for our experiments, we further partitioned the train set into train (90\% of samples from the original train set) and validation (the remaining 10\%) sets. We fit all models on this train set, and used performance on the validation set to optimize hyperparameters. We performed online augmentation during training, the details of which can be found in Appendix \ref{subsec:augmentation}. For a detailed description of the available hyperparameters, see Appendix \ref{subsec:hyperparameters}.

We considered three families of CNN backbone: DenseNet (DenseNet-121, DenseNet-161) \cite{densenet}, ResNet (ResNet-34, ResNet-50) \cite{resnet}, and VGG (VGG-16, VGG-19) \cite{vgg}. All backbones are pretrained on ImageNet \cite{imagenet} except for ResNet-50, which is the iNaturalist \cite{inaturalist} pretrained model of \cite{nauta2021neural}. We ran Bayesian hyperparameter tuning over models fit using a single NVIDIA RTX A5000 GPU on a private compute cluster, training four models in parallel, and limiting tuning to 12 computational days (3 calendar days) for all model constructions.

To distinguish the contributions of prototype similarity measures and prototype deformations, we optimize three different ProtoPNet variants implemented in the ProtoPNeXt framework: (1) The original ProtoPNet \cite{chen2019looks}, with a rigid prototype layer and Euclidean similarity between prototypes and images; (2) A rigid prototype layer with cosine similarity between the prototypes and images; (3) A deformable prototype layer implemented in slight variation from \cite{donnelly2022deformable} and cosine similarity (see: Appendix \ref{sec:deformable-impl}).
In our experiments, \textbf{we show that state-of-the-art accuracy can be achieved with a simple change in similarity metric and systematic hyperparameter tuning}. 

\textbf{Cosine Similarity Provides Superior Performance to Euclidean Distance.}
\begin{figure}[t]
    \centering
    \includegraphics[width=\textwidth]{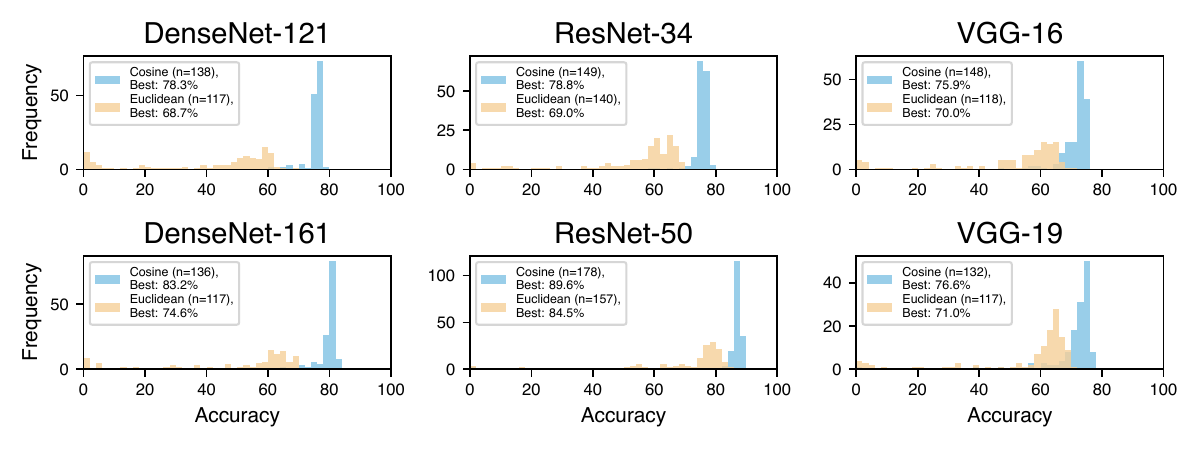}
    \caption{\textbf{Comparison of Trainability of Euclidean vs. Cosine Similarity Models.} The distribution of observed validation accuracy across all runs in 12 computational days of hyperparameter tuning. Using cosine similarity, we found a substantially higher optimal accuracy, a larger number of trained models, and a larger proportion of models trained achieved high accuracy.}
    \label{fig:l2_vs_cos}
\end{figure}
\begin{figure}[h]
    \centering
    \includegraphics[width=\textwidth]{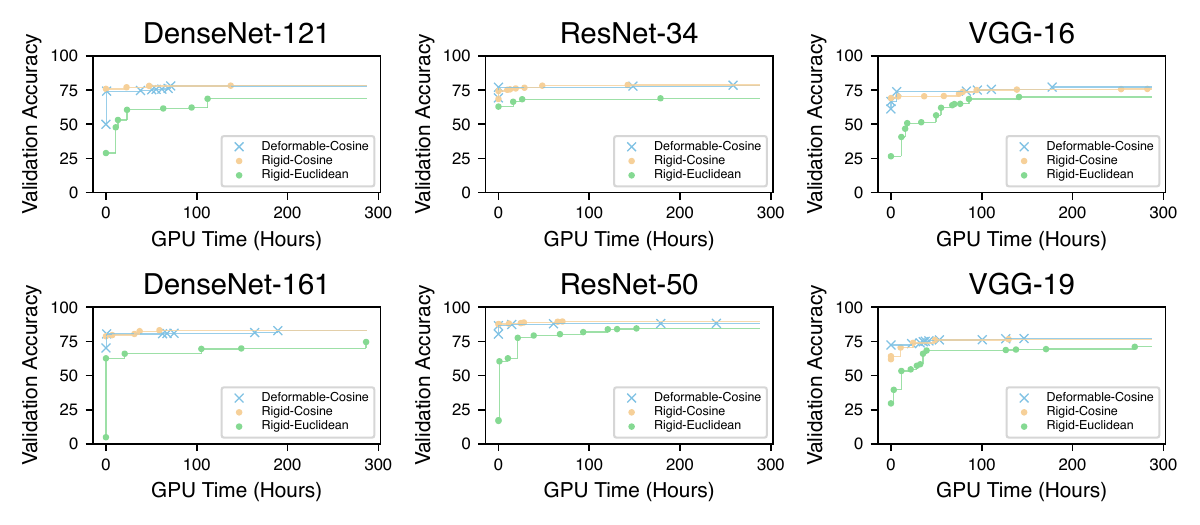}
    \caption{\textbf{Accuracy Progression by GPU-Hours.}
    GPU-hours are calculated as the product of the number of GPUs used (4) and the number of hours of training.
    The two cosine similarity models, `ProtoPNet with cosine' and `deformable', start with better performance and achieve saturation faster than `ProtoPNet with Euclidean distance.'
    `ProtoPNet with cosine distance' achieves saturation in under 50 GPU hours on Densenet-121 and VGG-16, and under 100 GPU hours on other backbones.}
    \label{fig:acc_vs_time}
\end{figure}
In prototypical-part models, prototypes are compared to image embeddings using either a Euclidean distance metric or a cosine similarity metric. Figure \ref{fig:l2_vs_cos} compares the test accuracy distribution between models implemented with these two metrics. We find that, across all six backbones, \textbf{models using cosine similarity tend to be more accurate than those using Euclidean distance}.
This trend holds both in terms of the whole distribution and in terms of the best observed model.
Figure \ref{fig:acc_vs_time} shows that models with cosine similarity also reach saturation during optimization more quickly than models with Euclidean distance.
In fact, models using cosine similarity and the initially selected hyperparameters in the optimization outperform the even best Euclidean models for all six of our experimental settings.
This suggests that \textbf{a large part of the superior accuracy of models such as Deformable ProtoPNet \cite{donnelly2022deformable} and TesNet \cite{wang2021interpretable} can be attributed to their use of cosine similarity}.

\textbf{Hyperparameter Tuning with Cosine Similarity is Sufficient for State of the Art Accuracy.}
\label{subsec:best-acc}
% \subsection{Experiment Setup}
% \label{subsec:experiment-setup}
% \TODO{Put anything here that doesn't belong in the Methods section. ie, GPU Config, runtime, weights and biases?}
% \subsection{Results}
% \label{subsec:acc-results}
Table \ref{tab:cub200-acc} reports the best test accuracy achieved by ProtoPNeXt for each backbone under the accuracy-only optimization objective. We find state-of-the-art accuracy on two of six backbones, \textbf{including the highest accuracy across all backbones} using ProtoPNet with only the change from Euclidean distance to cosine similarity, and competitive performance on all backbones under this fixed computation optimization. We repeat these experiments on the Stanford Dogs dataset \cite{KhoslaYaoJayadevaprakashFeiFei_FGVC2011}, and report the results in Appendix \ref{sec:stanford-dogs}, finding that the competitive performance persists.

% \begin{enumerate}
%     \item Provide a table of performance across backbones compared to prior work, really sell the numbers
%     \item Discuss the key hyperparameters (if interesting), e.g.``In our analysis, we encountered 3 surprising findings:
%     \begin{enumerate}
%         \item Deformations do not tend to provide performance benefits relative to non-deformed prototypes. This is surprising, given the strong performance of Deformable ProtoPNet. While this does not account for any interpretability benefits of deformable prototypes, it suggests that the performance of Deformable ProtoPNet may not be due to deformations themselves. Rather, we believe this performance may be due to the following point.
%         \item Cosine similarity provides a substantial improvement in performance relative to L2 distance. It is notable that two of the most performant case-based models -- Deformable ProtoPNet \cite{donnelly2022deformable} and TesNet \cite{rymarczyk2022interpretable} -- use cosine similarity for prototype comparisons.
%         \item Finally, prototypes consisting of one part consistently outperformed those consisting of multiple. ''
%     \end{enumerate}
% \end{enumerate}

\begin{table}[th]
    \centering
    \begin{tabular}{rcccccc}
\toprule
 & \multicolumn{2}{c}{DenseNet} & \multicolumn{2}{c}{ResNet}  & \multicolumn{2}{c}{VGG} \\
\textit{Model} & DN-121 & DN-161 & RN-34 & RN-50 & VGG-16 & VGG-19 \\
\midrule
\midrule
Baseline & 70.9 & 71.3 & 76.0 & 78.7 & 78.2 & 80.0 \\
ProtoPNet & 70.3 & 72.6 & 72.4 & 81.1 & 74.0 & 75.4 \\
Deformable ProtoPNet & \textbf{76.0} & \textbf{76.1} & \textbf{76.8} & 86.4 & \textbf{78.9} & 80.8 \\
ProtoTree &  &  &  & 82.2 &  &  \\
ProtoPNeXt - Best Test & 75.1 & 75.9 & 75.3 & \textit{\textbf{87.0}} & 77.6 & \textit{\textbf{80.9}} \\
ProtoPNeXt - Best Val & 74.4 & 74.1 & 74.4 & 86.4 & 75.5 & 80.7 \\
\bottomrule
\end{tabular}

    \caption{\textbf{Uncropped CUB-200 Test Accuracy.} Comparison of each model across six CNN backbones.
    The Best Test model is selected by best model performance on the test set, irrespective of performance on the validation set. This is comparable to other works that do not include a separate validation split in their training. The Best Val model is selected by accuracy on the validation set. \textbf{bold} -- best accuracy  for backbone. \textit{italics} -- uncropped CUB-200 state of the art for prototypical-part models.}
    \label{tab:cub200-acc}
\end{table}

%In particular, by %tuning only on hyperparameters of ProtoPNet \cite{chen2019looks} and Deformable ProtoPNet \cite{donnelly2022deformable}, we can achieve an optimized accuracy of PLACEHOLDER\%, an improvement of PLACEHOLDER\% over models provided by the original paper's team.

%The codebase for ProtoPNext has integrated API calls for \textit{Weights \& Biases}, which provides a Bayesian hyperparameter tuning ... Using this, tuning only on hyperparameters mention in the original ProtoPNet paper, we can achieve an optimized accuracy of __\%, an improvement of __\% over models provided by the original paper's team.

\textbf{Deformable Prototypes Tend Not to Improve Performance, but Generalize Well.}
\begin{table}
    \centering
    \begin{tabular}{rrcccccc}
\toprule
 \multirow{2}*{\textit{Prototypes}} & \multirow{2}*{\textit{Selection}}& \multicolumn{2}{c}{DenseNet} & \multicolumn{2}{c}{ResNet}  & \multicolumn{2}{c}{VGG} \\
 &  & DN-121 & DN-161 & RN-34 & RN-50 & VGG-16 & VGG-19 \\
\midrule
\multirow{2}{*}{Rigid}
 & Best Val & 75.5 & 80.7 & 74.4 & 86.4 & 74.2 & 74.1 \\
 & Best Test  & 77.6 & 80.9 & 75.3 & 87.0 & 74.7 & 74.9 \\
\midrule
\multirow{2}{*}{Deformable} &  Best Val  & 75.4 & 79.1 & 73.6 & 85.5 & 74.4 & 74.0 \\
 &  Best Test & 77.0 & 79.9 & 75.1 & 86.8 & 75.1 & 75.9 \\
\bottomrule
\end{tabular}

    \caption{\textbf{Comparing Fixed-Cost Performance of Similarity Measures and Prototype Layers.} Comparison of ProtoPNeXt test accuracy on uncropped CUB-200 with deformable and non-deformable prototypes. We performed separate fixed-cost optimizations by fixing the prototype layer -- rigid vs$.$ deformable prototypes -- and the similarity metric to cosine. \textit{Selection} is whether the model was selected for its validation accuracy or test accuracy.}
    \label{tab:cub200-acc-deform-comp}
\end{table}
Table \ref{tab:cub200-acc-deform-comp} presents the best observed test accuracy when using deformable and non-deformable prototypes. We observe little difference between models with deformable prototypes and non-deformable prototypes under this fixed GPU-cost optimization scheme despite the added expressiveness of deformable prototypes.
The absence of performance improvement may be the result of the complexity of optimizing models with deformable prototypes, which have more hyperparameters and took longer to achieve their best performance during hyperparameter sweeps (Figure \ref{fig:acc_vs_time}).
Conversely, deformable prototypes show marginally better generalization of accuracy from the validation set to the test set: mean $-0.9$\% for deformable prototypes, mean $-1.3$\% for non-deformable prototypes using cosine distance ($p<0.001$ using a one-sided t-test with $1713$ runs; see distribution in Appendix, Figure \ref{fig:acc-generalization}).

\section{Optimizing for Interpretability}
\label{sec:interp-experiments}
Having shown that simply applying cosine similarity and carefully tuning hyperparameters is sufficient to achieve the recent performance gains in the literature, we now turn to investigate whether improvements in model interpretability can be similarly achieved.
As before, for each backbone CNN considered, we ran Bayesian hyperparameter optimization for 12 computational days, restricted to models using cosine similarity.
In these settings, we maximize the joint objective from Equation \ref{eq:joint_obj}.

\textbf{Joint Optimization Yields Better Prototypes Without Sacrificing Accuracy.}
We computed the accuracy, stability, consistency, and sparsity of the prototypes of models trained during the accuracy-only optimization and this new joint optimization, and compared these metrics.
Table \ref{tab:cub200-interp} shows the results of this evaluation. We find that, \textbf{across all backbones, jointly optimizing hyperparameters for accuracy and prototype quality produces models with improved prototype quality without sacrificing accuracy}. That is, joint optimization allows us to create models that are easier to interpret with no additional computational or performance cost. Appendix \ref{subsec:app-corr-acc-interp} shows that there are many equally accurate models with very different prototype quality scores.

\begin{table}
    \centering
    \begin{tabular}{rrcccccc}
\toprule
 \multirow{2}*{\textit{Prototypes}} & \multirow{2}*{\textit{Optimization Obj.}} & \multicolumn{2}{r}{DenseNet-161} & \multicolumn{2}{r}{ResNet-50} & \multicolumn{2}{r}{VGG-19} \\
  & & Acc & PS & Acc & PS & Acc & PS \\
\midrule
\multirow{3}{*}{Rigid} & Accuracy & 80.7 & 60.0 & 86.4 & 67.6 & 74.1 & 50.6  \\
 & Accuracy-Prototype Score & 79.2 & 73.6 & 86.2 & 81.4 & 73.5 & 63.6 \\
 & Difference & -1.5 & +13.6 & -0.2 & +13.8 & -0.6 & +13.0 \\
\cline{1-8}
\multirow{3}{*}{Deformable} & Accuracy & 79.1 & 40.0 & 85.5 & 39.6 & 74.0 & 17.9  \\
 & Accuracy-Prototype Score & 79.5 & 42.8 & 84.8 & 49.6 & 75.3 & 33.8  \\
 & Difference & +0.4 & +2.8 & -0.7 & +10.0 & +1.3 & +15.9  \\
\cline{1-8}
\bottomrule
\end{tabular}

    \caption{\textbf{Joint optimization substantially improves interpretability metrics without substantial cost to accuracy.} \textit{Acc} is accuracy. \textit{PS} is Prototype Score. Models were selected for their validation accuracy, and performance is reported on the test set. From joint optimization, across backbones, we see improvement on prototype scores with a range of +2.8 to +15.9 and a change of -1.5\% to +1.3\% in accuracy. Both metrics range from 0 - 100. Appendix \ref{sec:extended-results} shows the same comparisons for models selected by test accuracy and by the joint accuracy-prototype objective, with similar results.}
    \label{tab:cub200-interp}
\end{table}
\begin{figure}
    \centering
     \includegraphics[width=0.96\textwidth]{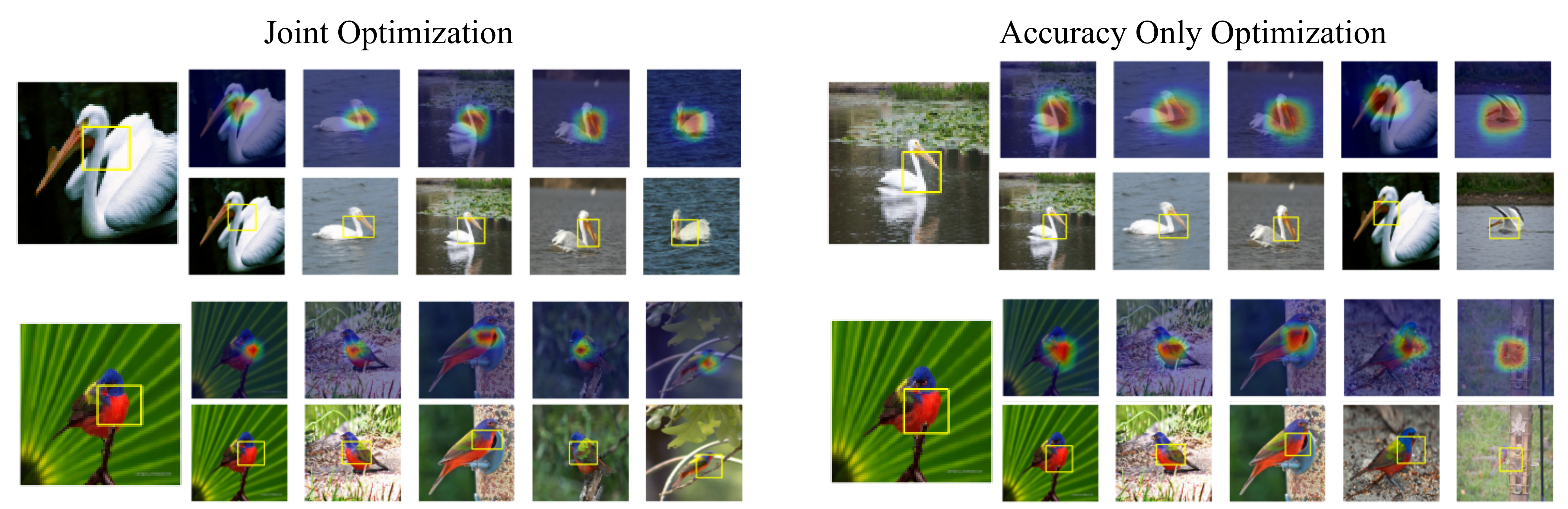}
    \caption{\textbf{Comparing global analysis of best joint- and accuracy-only-optimized models.} The leftmost image in each collection is a prototype, followed by the five images with the highest activations for that prototype. Models were selected for best validation accuracy across all configurations. Prototypes from the jointly optimized model are more precise and consistent. Joint model: 86.2\% test accuracy, 81.4 test prototype score; Accuracy only: 86.4\% test accuracy, 67.6 prototype score.}
    \label{fig:global-analysis}
\end{figure}

\textbf{Joint Optimization Yields Qualitatively Better Models.}
Here, we demonstrate that models with a high interpretability score according to Equation \ref{eq:proto_score} have substantially stronger semantics. %To compare the semantics of models with and without strong interpretability scores, 
We analyze the most accurate model produced by optimizing for accuracy alone, and the model with the highest accuracy produced by joint optimization. 

We evaluate whether the prototypes from each model reliably represent a single, semantically meaningful part. Figure \ref{fig:global-analysis} presents the 5 most similar images to two prototypes from each model. We see that prototypes from the jointly optimized model show greater semantic consistency and precision, with tight, consistent activation regions.  
% \TODO{Add discussion of the global analysis fig, highlighting how nice the high metric vis is} 
The full set of prototypes for each model, example reasoning processes, and additional global analyses are available in Appendix \ref{sec:more-proto-examples}.

\section{Conclusion}
\label{sec:conclusion}
We studied the source of recent improvements in case-based DNN performance. To enable this, we introduced ProtoPNeXt, a framework for easily creating and optimizing case-based DNN's. We introduced a novel interpretability metric and stopping criteria tailored for case-based DNN's, which allowed us to systematically apply hyperparameter tuning to study the source of accuracy and interpretability improvements in the literature. We used this approach to train models with superior accuracy, showing that cosine similarity is responsible for much of the performance improvement seen in recent case-based DNN's. Further, we showed that jointly optimizing for interpretability and accuracy can explain improved model semantics - in our study, these are models that are substantially sparser, more consistent, and more stable without sacrificing accuracy.

This work has substantial implications for future research on case-based DNNs and for their practical use. 
For researchers, our results suggest that future case-based DNNs should 1) default to cosine similarity 2) use case-based models that are directly optimized for interpretability and 3) leverage systematic hyperparameter tuning. In particular, the effectiveness of joint optimization suggests the need for further research into prototype quality metrics to systematize measurement of prototype quality.
For practical applications, the ProtoPNeXt framework allows easy hyperparameter tuning along multiple objectives for new datasets, reducing the difficulty in applying these models.
ProtoPNeXt has been carefully developed for modularity, and the code will be released upon publication.
This will support future work expanding the framework to include features from a broad set of case-based DNNs.

\paragraph{Limitations.}
\label{sec:limitations}
This work is limited by its scope; there are many extensions to ProtoPNet, and an optimization over case-based DNNs should incorporate all of them. As such, the analyses in this paper do not yet show the full potential of this line of work.

The goal of optimizing prototype quality metrics is to achieve more interpretable models.
We showed that different models with the same architecture and similar predictive performance may have different prototype quality metrics,
but we are limited in our optimizations by how well these metrics measure interpretability.
There are other desirable qualities of prototypes 
%- for instance, saliency of a feature to human observers - 
that are not accounted for in these metrics.
As such, we cannot say that our optimization objective represents an ideal mix of prototype qualities, only that it expresses some desirable qualities.
The even weighting of our optimization was chosen because of uncertainty of what optimized values were achievable, but it may not represent an appropriate weighting for any given task.
A complete study of prototype quality metrics is beyond the scope of this work.

In our experiments, we did not choose to optimize the latent space size in our hyperparameter tuning, which directly affects how much of the input image each prototypical part represents.
Instead, we used a common size of 7x7.
Since we experimented on uncropped CUB-200 images, for images where the bird only appears in a small portion, the resulting prototypical parts can cover large areas of the bird. A more comprehensive study would also tune the latent space size.

\paragraph{Societal Impact.}
\label{sec:societal}
The question of how to make deep learning algorithms for computer vision interpretable is one of the most important considerations for trust in AI. The answers to this could heavily impact self-driving cars, radiology, and facial recognition in policing.

% \section{Author Contribution Statement}
% \label{sec:author_contribution_statement}
% \input{author_cont}

\bibliography{references}
\bibliographystyle{ieee_fullname}

\appendix

\section{Detailed Experimental Setup}
\label{sec:detailed-setup}
\subsection{Bayesian Hyperparameter Optimization}
\label{subsec:bayesian-optimization}

Hyperparameter selection is not well documented in the literature even though these models introduce a number of key hyperparameters that control the balance between optimizing performance and interpretability. Because hyperparameter tuning requires many expensive model fits, we leverage Bayesian hyperparameter tuning. Bayesian hyperparameter estimates a posterior distribution over hyperparameter configurations with a cheap-to-evaluate surrogate function. Using this estimate, the optimization regime finds a set of hyperparameters over which to perform a complete training run that will resolve uncertainty while having a high likelihood of success. For a more complete review of Bayesian hyperparameter tuning, see \cite{bischl2023hyperparameter}. In practice, we use the optimization regime implemented by Weights and Biases Sweeps \cite{wandb} under Academic Researching Licensing. The specific hyperparameters we optimize are provided in section \ref{subsec:hyperparameters}.

\subsection{Hyperparameters}
\label{subsec:hyperparameters}

Below, we describe the specific hyperparameters tuned through our Bayesian hyperparameter tuning. 
For all runs, we tune the following hyperparameters:

\begin{itemize}
    \item \textit{pre\_project\_phase\_len} (discrete; min 3, max 15) --- Number of epochs in each pre-project training phase (warm-up, joint). A value of 3 for this (with a phase\_multiplier of 1) means that there will be 3 warm-up epochs and 3 joint epochs before the first model project step. Total preproject epochs is 2 * pre\_project\_phase\_len * phase\_multiplier.
    \item \textit{post\_project\_phases} (discrete; min 10, max 10) --- Number of times to iterate between last-only, joint, project after the initial pre-project phases. Our fixed value of 10 means that there will be 10 projects within the joint training (with each project step followed by last-only epochs)
    \item \textit{phase\_multiplier} (discrete; min 1, max 1) --- For each phase, multiply the number of epochs in that phase by this number. Our fixed value of 1 means that we will not alter the number of epochs.
    \item \textit{num\_addon\_layers} (discrete; min 0, max 2) --- Number of optional add-on layers to include in the model, where add-on layers lie between the model backbone and prototype layer. A value of 0 means that the backbone is connected directly to the prototype layer, whereas a value of 1 or 2 will add num\_addon\_layers convolutional layers (with ReLU), ending with a Sigmoid activation function.
    \item \textit{latent\_dim\_multiplier\_exp} (discrete; min -4, max 1) ---  Exponential of 2 for the latent dimension of the prototype layer. Will be 0 if there are 0 add-on layers. If there are add-on layers, this will be multiplied by the number of input channels to determine the number of output channels for the convolutions in the add-on layers.
    \item \textit{num\_prototypes\_per\_class} (discrete; min 1, max 16) --- Number of prototypes per class.
    \item \textit{joint\_lr\_step\_size} (discrete; min 2, max 10)  --- Number of epochs between each step in the joint learning rate scheduler. Multiplied by phase\_multiplier (fixed at 1).
    \item \textit{lr\_multiplier} (Normal; $\mu=1.0, \sigma=0.4$) --- Multiplier for learning rates (same for all training phases of warm-up, joint, last-layer). Used to jointly tune the learning rates
    \item \textit{cluster\_coef} (Normal; $\mu=-0.8, \sigma=0.5$) --- Coefficient for clustering term in loss function. Our clustering term encourages a segment of the latent representation of an image from a particular class is near at least one prototype of that class. This ensures that the prototypes "cluster" around the features characteristic of the class they represent and that images within the same class are closer in the latent space.
    \item \textit{separation\_coef} (Normal; $\mu=0.08, \sigma=0.1$) --- Coefficient for separation term in loss function. The separation term attempts to maximize the smallest distance between prototypes not of the same class as a training image and the latent representation of the training image. This ensures that prototypes that do not belong to a given class are distant from the latent space of the other class, encouraging differences within the class.
    \item \textit{l1\_coef} (Log Uniform; min=0.00001, max=0.001) --- Coefficient for L1 Regularization of model in loss function.\\
\end{itemize}

When tuning deformable prototypes, we include additional hyperparameters relating to features from Deformable ProtoPNet. In particular, we add:
\begin{itemize}
    \item \textit{num\_warm\_pre\_offset\_epochs} (discrete; min 0, max 10) --- Number of epochs spent optimizing prototypes and add-on layers, but not offsets or the backbone.
    \item \textit{k\_for\_topk} (discrete; min 1, max 10) --- The number of prototype activation locations to average over when computing prototype similarity.
    \item \textit{prototype\_dimension} (discrete; min 1, max 3) --- The spatial size of each prototype; a prototype with \textit{prototype\_dimension}=3 has 9 parts.
    \item \textit{orthogonality\_loss} (Log uniform; min 0.00001, max 0.001) --- The coefficient applied to the orthogonality loss from Deformable ProtoPNet.
\end{itemize}

\subsection{Augmentation Details}
\label{subsec:augmentation}
Throughout our experiments, we perform online augmentation on our training data to prevent overfitting. In particular, when loading each training image, we performed:
\begin{itemize}
    \item A random rotation between -15 and 15 degrees
    \item A random distortion with scale 0.2
    \item A random shear of up to 10 pixels
    \item A 50\% chance of a horizontal flip
\end{itemize}

\section{Deformable Prototype Implementation}
\label{sec:deformable-impl}
In Deformable ProtoPNet \cite{donnelly2022deformable}, each prototype and each latent feature vector is restricted to be of a fixed norm. In particular, if prototypes consist of $H_P W_P$ parts, each part and each latent vector is set to have a 2-norm of $\frac{1}{\sqrt{H_PW_P}}.$ This guarantees that the cosine similarity between a prototype $\mathbf{p}_j \in \mathbb{R}^{d \times H_P \times W_P}$ and a group of latent feature vectors $\{\mathbf{z}_k \in \mathbb{R}^d\}_{k=1}^{H_P W_P}$ can be computed directly as:
\begin{align*}
\text{cos}\left(\theta\left(\begin{bmatrix}
        \mathbf{p}_{j, :, 1, 1}\\
        \mathbf{p}_{j, :, 1, 2}\\
        \hdots\\
        \mathbf{p}_{j, :, H_P, W_P}\\
    \end{bmatrix}, 
    \begin{bmatrix}
        \mathbf{z}_{1}\\
        \mathbf{z}_{2}\\
        \hdots\\
        \mathbf{z}_{H_PW_P}\\
    \end{bmatrix}\right)\right) &= 
    \begin{bmatrix}
        \mathbf{p}_{j, :, 1, 1}\\
        \mathbf{p}_{j, :, 1, 2}\\
        \hdots\\
        \mathbf{p}_{j, :, H_P, W_P}\\
    \end{bmatrix}^T
    \begin{bmatrix}
        \mathbf{z}_{1}\\
        \mathbf{z}_{2}\\
        \hdots\\
        \mathbf{z}_{H_PW_P}\\
    \end{bmatrix} = \sum_{h}^{H_P} \sum_{w}^{W_P} \langle\mathbf{p}_{j, :, h, w}, \mathbf{z}_{hw}\rangle,
\end{align*}
where $\theta$ denotes the angle between two vectors.

Deformations complicate this formulation, since the feature at a fractional location is defined to be an interpolation between the neighboring grid locations. In general, bilinear interpolation does not preserve the norm of the vectors it interpolates between. Deformable ProtoPNet \cite{donnelly2022deformable} introduces a norm-preserving interpolation method to address this issue, but implementing this interpolation function requires complicated, low-level changes to a standard implementation of deformable convolution. In ProtoPNext, we opted instead to use standard PyTorch \cite{paszke2019pytorch} functionality rather than a custom implementation of deformable convolution. This offers two benefits: this approach is less error-prone since the PyTorch functionality is likely to receive support moving forward, and this approach avoids the need to build and install custom C++ code, as is required with the implementation from \cite{donnelly2022deformable}. 

Concretely, in our deformable prototype implementation we 
\begin{enumerate}
    \item Predict a set of offsets using convolution, as in \cite{donnelly2022deformable}.
    \item Compute the feature value at each location suggested by these offsets using bilinear interpolation. This is implemented using PyTorch's version of the sampling method from \cite{jaderberg2015spatial}.
    \item Re-normalize each interpolated vector to have the desired norm.
    \item Compare our prototypes to these interpolated vectors, getting a similarity score at each required location.
\end{enumerate}

Steps 2, 3, and 4 differ from the implementation suggested in \cite{donnelly2022deformable}, but maintain all aspects of the method other than the interpolation function used. 

\section{Extended Optimization Results}
\label{sec:extended-results}
\subsection{Accuracy-Only Optimizations}

\begin{figure}[t]
    \centering
    \includegraphics[width=\textwidth]{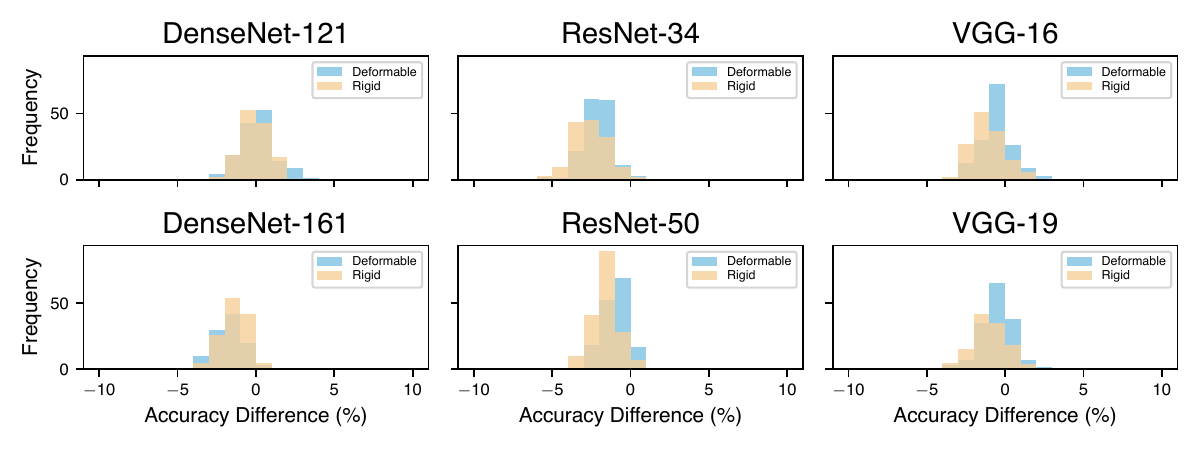}
    \caption{\textbf{Comparing Generalization of Rigid Prototypes to Deformable Prototypes.} Accuracy Difference is the difference between validation accuracy and test accuracy. Deformable prototypes have marginally better generalization than rigid using cosine similarity.}
    \label{fig:acc-generalization}
\end{figure}

Figure \ref{fig:acc-generalization} shows the distributions of differences between validation set accuracy and test set accuracy on per model and backbone.
In our accuracy-only optimizations, deformable prototypes generalize better than single prototypes.

\subsection{Distribution of Accuracy and Prototype Metrics}
\label{subsec:app-corr-acc-interp}

\begin{figure}[t]
    \centering
    \includegraphics[width=\textwidth]{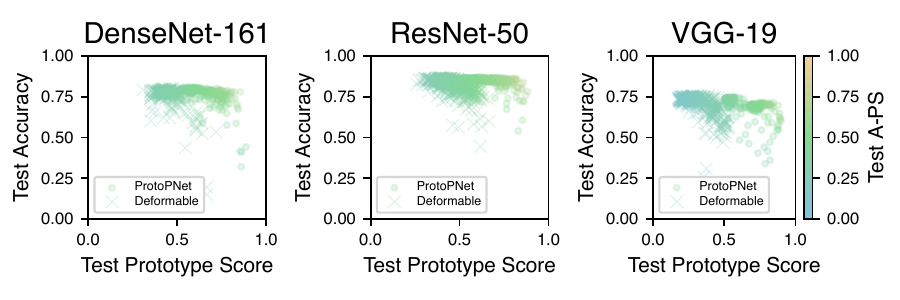}
    \caption{\textbf{Model Accuracy and Prototype Score Correlation.} \textit{A-PS} is accuracy-protype score, the joint optimazation $\text{obj}_{\text{aps}}.$
    Includes ProtoPNet and Deformable ProtoPNet models with Cosine similarity optimized under both accuracy and accuracy-prototype score objectives.
    There are a large number of models that exhibit similar accuracy but have a broad spectrum of prototype quality scores.
    This provides an empirical justification for how joint optimization can improve prototype quality without sacrificing accuracy.}
    \label{fig:acc-interp-correlation}
\end{figure}

In the main paper, we showed that optimizing for both accuracy and a prototype score does not reduce the accuracy of the resulting model. Figure \ref{fig:acc-interp-correlation} shows that, across both the accuracy-only optimization and the joint accuracy-prototype score optimization, there are a large number of models that have similar accuracies but materially different prototypes quality scores, which provides an empirical justification for how improvement is possible.

\subsection{Joint Accuracy-Prototype Score Optimizations}
In the body of the paper, we compared joint accuracy-prototype quality optimization results to accuracy-only optimization using model selection on best validation accuracy.
In this section, we report the same results using the following model selection criteria:

\begin{enumerate}
    \item Test Accuracy (Table \ref{tab:cub200-interp-test})
    \item Validation Accuracy-Prototype Score (Table \ref{tab:cub200-interp-aps})
    \item Test Accuracy-Prototype Score (Table \ref{tab:cub200-interp-aps})
\end{enumerate}

It is important to note that the model selection criteria is a decision about which model properties are most desirable.
In general, practitioners will not have access to test results during selection.

\begin{table}
    \centering
    \begin{tabular}{rrrcccccc}
\toprule
  \multirow{2}*{\textit{Selection}} &
 \multirow{2}*{\textit{Prototypes}} & 
 \multirow{2}*{\textit{Optimization Obj.}} & \multicolumn{2}{r}{DenseNet-161} & \multicolumn{2}{r}{ResNet-50} & \multicolumn{2}{r}{VGG-19} \\
 &  &  & Acc & PS & Acc & PS & Acc & PS \\
\midrule
\multirow[c]{4}{*}{Best Val} & \multirow[c]{2}{*}{Rigid} & Acc  & 80.7 & 60.0 & 86.4 & 67.6 & 74.1 & 50.6 \\
 &  & A-PS & 79.2 & 73.6 & 86.2 & 81.4 & 73.5 & 63.6 \\
\cline{2-9}
 & \multirow[c]{2}{*}{Deformable} & Acc & 79.1 & 40.0 & 85.5 & 39.6 & 74.0 & 17.9  \\
 &  & A-PS & 79.5 & 42.8 & 84.8 & 49.6 & 75.3 & 33.8  \\
\cline{1-9} \cline{2-9}
\multirow[c]{4}{*}{Best Test} & \multirow[c]{2}{*}{Rigid} & Acc & 80.9 & 56.0 & 87.0 & 61.1 & 74.9 & 56.9 \\
 &  & A-PS & 80.2 & 72.3 & 87.0 & 72.3 & 73.8 & 61.2 \\
\cline{2-9}
 & \multirow[c]{2}{*}{Deformable} & Acc & 79.9 & 42.3  & 86.8 & 38.2 & 75.9 & 38.7 \\
 &  & A-PS & 79.9 & 46.2 & 85.4 & 33.5 & 75.3 & 33.8  \\
\cline{1-9} \cline{2-9}
\bottomrule
\end{tabular}

    \caption{\textbf{Accuracy and interpretability metrics for the best joint optimized model versus those for the best accuracy optimized model selected by Accuracy.} \textit{A-PS} is Accuracy-Prototype Score, defined in Equation \ref{eq:proto_score}. \textit{Acc} is accuracy. \textit{PS} is Prototype Score. Metrics are from the model with best performance according to the sweep's objective, either $\text{obj}_{\text{acc}}$ or $\text{obj}_{\text{aps}}$. Models are organized by whether the were selected by best performance on test set (\texttt{test}) or the validation set (\texttt{val}).}
    \label{tab:cub200-interp-test}
\end{table}

\begin{figure}
    \centering
    \includegraphics[width=\textwidth]{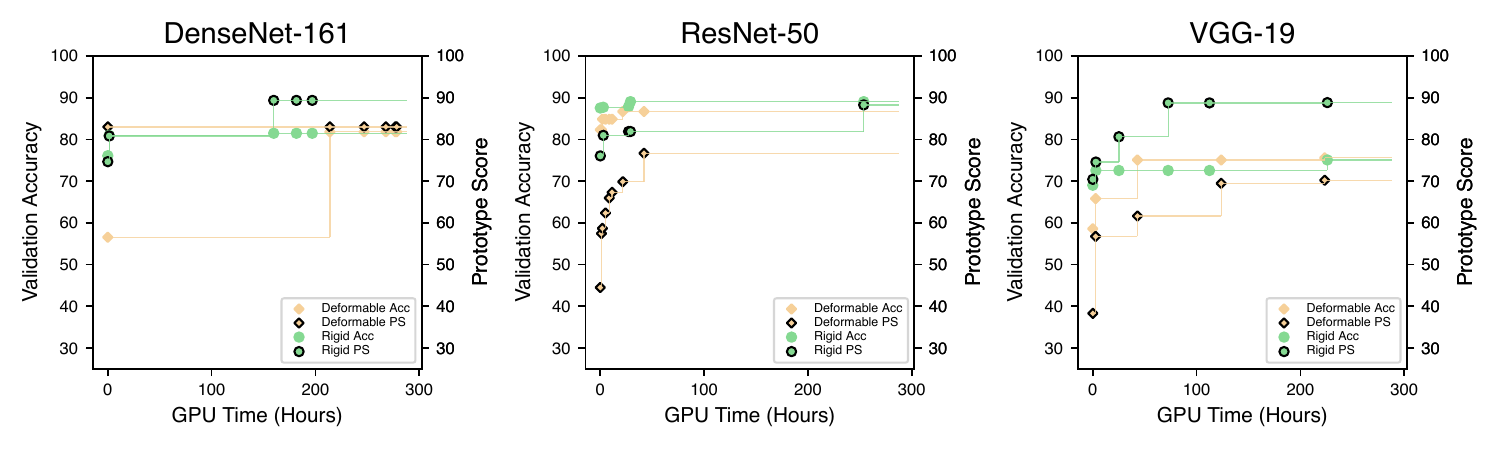}
    \caption{\textbf{Accuracy and Prototype Score Progression by GPU-Hours.}
    GPU-hours are calculated as the product of the number of GPUs used (4) and the number of hours of training. Each simultaneous two points for a particular model architecture (e.g., ProtoPNet) are the validation scores for the single model with the best $\text{obj}_{\text{aps}}$. It is notable that the two metrics increase jointly in the very early phases of optimization, but then separately thereafter, meaning that during optimization one metric is improving without harming the other. In comparison to Figure \ref{fig:acc_vs_time}, models do not consistently saturate on this joint objective during the early optimization window.}
    \label{fig:interp_vs_time}
\end{figure}

Figure \ref{fig:interp_vs_time} shows how model accuracy and prototype score progress during joint optimization.

\begin{table}
    \centering
    \begin{tabular}{rrrcccccc}
\toprule
\multirow{2}*{\textit{Selection}} &
 \multirow{2}*{\textit{Prototypes}} & 
 \multirow{2}*{\textit{Optimization Obj.}} & \multicolumn{2}{r}{DenseNet-161}  & \multicolumn{2}{r}{ResNet-50} & \multicolumn{2}{r}{VGG-19} \\
 &  &  & Acc & PS & Acc & PS & Acc & PS \\
\midrule
\multirow[c]{4}{*}{Best Val} & \multirow[c]{2}{*}{Rigid} & Acc & 78.3 & 69.0 & 82.2 & 80.9 & 67.4 & 75.4  \\
 &  & A-PS & 77.9 & 85.6 & 83.8 & 85.9 & 69.8 & 87.8  \\
\cline{2-9}
 & \multirow[c]{2}{*}{Deformable} & Acc & 77.4 & 40.1 & 81.4 & 60.4 & 73.6 & 39.4 \\
 &  & A-PS & 76.4 & 45.9 & 79.8 & 56.6 & 74.6 & 45.7 \\
\cline{1-9} \cline{2-9}
\multirow[c]{4}{*}{Best Test} & \multirow[c]{2}{*}{ProtoPNet} & Acc & 79.0 & 73.9 & 85.1 & 78.2 & 67.4 & 75.4 \\
 &  & A-PS & 77.9 & 85.6 & 83.8 & 85.9 & 69.8 & 87.8 \\
\cline{2-9}
 & \multirow[c]{2}{*}{Deformable} & Acc & 74.4 & 55.6 & 81.4 & 60.4 & 75.9 & 38.7 \\
 &  & A-PS & 74.8 & 58.6 & 79.4 & 62.2 & 74.6 & 45.7  \\
\cline{1-9} \cline{2-9}
\bottomrule
\end{tabular}

    \caption{\textbf{Accuracy and interpretability metrics for the best joint optimized model versus those for the best accuracy optimized model selected by Accuracy-Prototype Score.} \textit{A-PS} is Accuracy-Prototype Score, defined in Equation \ref{eq:proto_score}. \textit{Acc} is accuracy. \textit{PS} is Prototype Score. Metrics are from the model with best performance according to the sweep's objective, either $\text{obj}_{\text{acc}}$ or $\text{obj}_{\text{aps}}$. Models are organized by whether the were selected by best performance on test set (\texttt{test}) or the validation set (\texttt{val}).}
    \label{tab:cub200-interp-aps}
\end{table}

\section{Stanford Dogs}
\label{sec:stanford-dogs}
\begin{table}[]
    \centering
    \begin{tabular}{rrccc}
\toprule
\textit{Model} & \textit{Selection} & DenseNet-161 & ResNet-50 & VGG-19 \\
\midrule
Baseline & & 84.1 &  & 77.3 \\
ProtoPNet & & 77.3 &  & 73.6 \\
Deformable ProtoPNet & & 86.5 &  & 77.9 \\
\multirow{2}{*}{ProtoPNeXt -Rigid} & Best Val & 85.4 & 66.8 & 78.8 \\
& Best Test & 85.4 & 67.4 & 78.8 \\
\multirow{2}{*}{ProtoPNeXt -Deformable}  & Best Val & 81.8 & 68.1 & 74.3\\
 & Best Test & 82.3 & 68.6 &  75.0 \\

\bottomrule
\end{tabular}

    \caption{\textbf{Accuracy for Stanford Dogs.} Accuracies from accuracy-only optimization of ProtoPNeXt on Stanford Dogs dataset. Optimization was run for 6 computational days with a parallelism of 4 (1.5 wall days) on a single Nvidia A5000 for each run. Only cosine similarity models were trained. Deformable and Rigid are prototype layers.}
    \label{tab:dogs-acc}
\end{table}

In the main body of this paper, we focused on evaluations using the CUB-200 image classification dataset \cite{WahCUB_200_2011}. In this section, we turn to evaluate the Stanford Dogs dataset \cite{KhoslaYaoJayadevaprakashFeiFei_FGVC2011}.
The Stanford Dogs dataset provides standardized train and test splits; as with  CUB-200, we further partitioned the train set into train (90\% of samples from the original train set) and validation (the remaining 10\%) sets for our experiments. We fit all models on this train set, and used performance on the validation set to optimize hyperparameters.

We repeat the accuracy experiment from Section \ref{sec:experiments} on the Stanford Dogs dataset with the following changes: We limit the total computational time to 6 days, we test only models using cosine similarity, and we study one backbone from each family: DenseNet-161, ResNet-50, and VGG-19. Table \ref{tab:dogs-acc} reports the accuracy achieved by the ProtoPNeXt framework across the VGG-19, ResNet-50, and DenseNet-161 backbones. We find that, as with CUB-200, ProtoPNeXt tends to produce comparable accuracy to prior methods.

\clearpage
\section{Additional Visualizations}
\label{sec:more-proto-examples}
In this section, we provide a large set of additional  visualizations from the model jointly trained for accuracy and interpretability that was discussed in Section \ref{sec:interp-experiments}. 

\begin{figure}
    \centering
    \includegraphics[width=\linewidth]{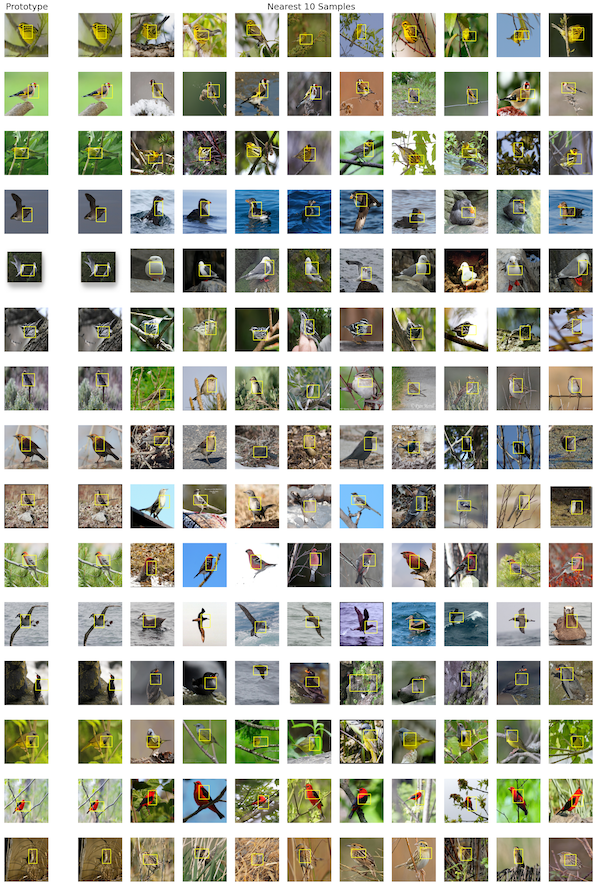}
    \caption{Each row in the figure shows the prototype in the first column, followed by the ten images with the highest activations for that prototype (sorted by activation in descending order). In each image, a yellow bounding box is shown to highlight the image patch match to the prototype. The closest match to the prototype is always itself.}
    \label{fig:global_analysis_examples}
\end{figure}

Figure \ref{fig:global_analysis_examples} illustrates 15 prototypes from this model and the 10 input images that yielded the highest activation for them. Across all 15 prototypes, we see strong, consistent semantics in what the prototypes look for, further showing that jointly optimizing for prototype metrics improves the semantics of the resulting model.

\begin{figure}
    \centering
    \includegraphics[width=\textwidth]{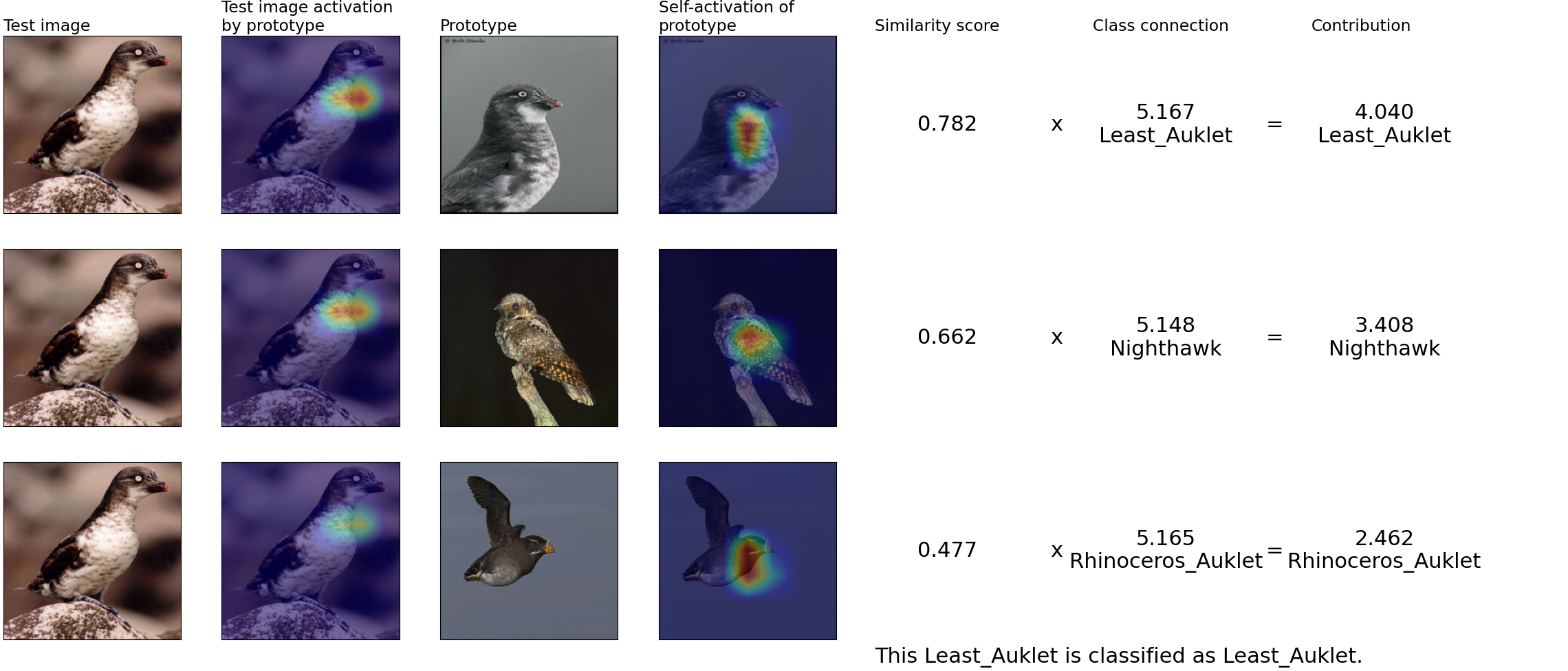}
    \caption{Reasoning process for the best model trained to jointly optimize accuracy and interpretability on an image of a Least Auklet. The model correctly classifies this image.}
    \label{fig:local_17}
\end{figure}

\begin{figure}
    \centering
    \includegraphics[width=\textwidth]{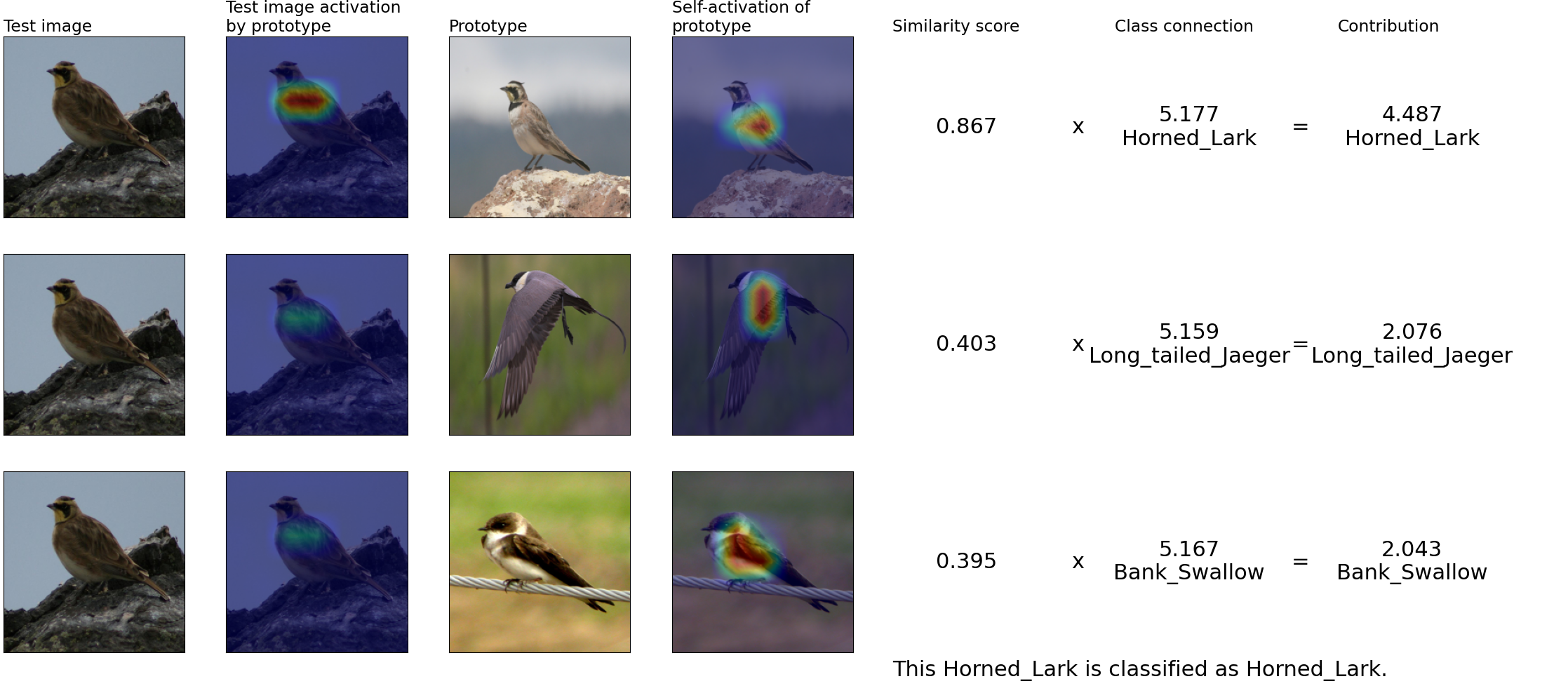}
    \caption{Reasoning process for the best model trained to jointly optimize accuracy and interpretability on an image of a Horned Lark. The model correctly classifies this image.}
    \label{fig:local_253}
\end{figure}

\begin{figure}
    \centering
    \includegraphics[width=\textwidth]{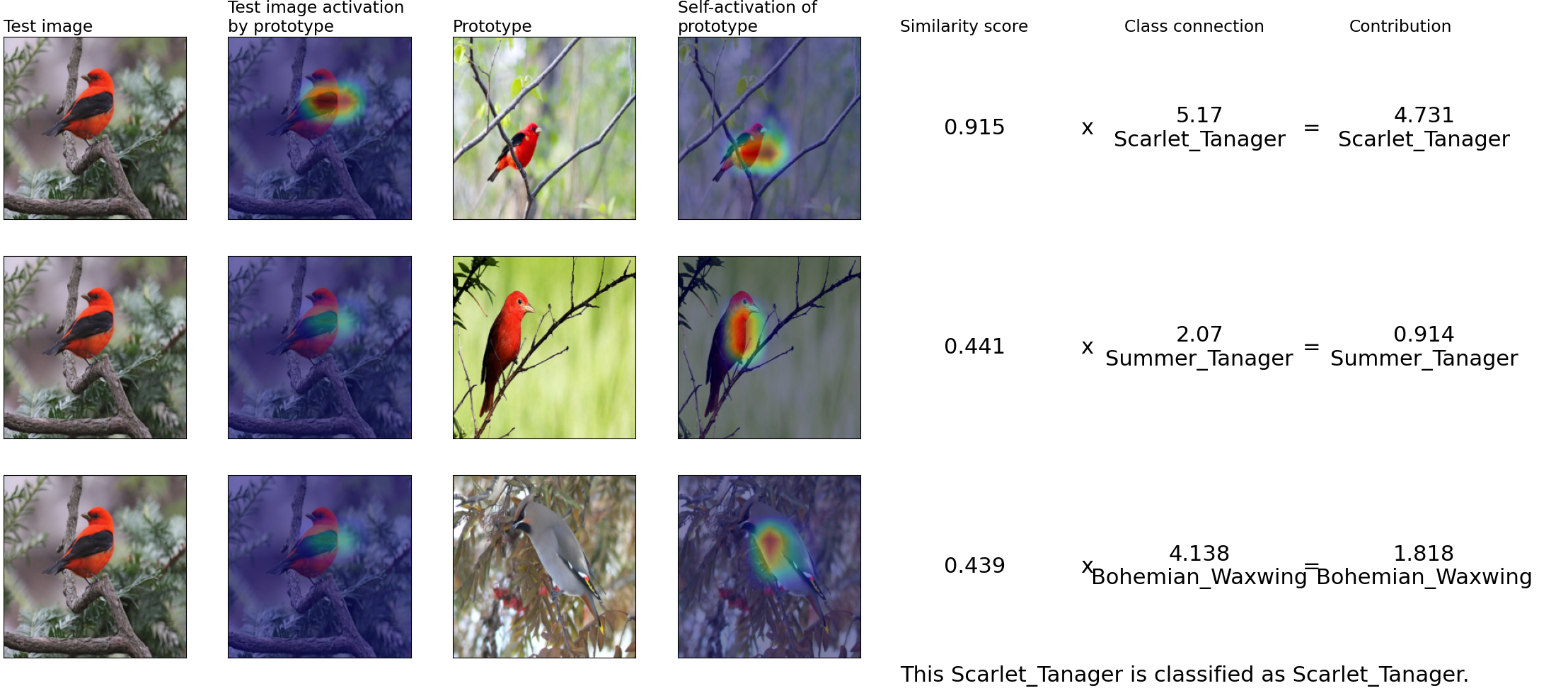}
    \caption{Reasoning process for the best model trained to jointly optimize accuracy and interpretability on an image of a Scarlet Tanager. The model correctly classifies this image.}
    \label{fig:local_411}
\end{figure}

Additionally, several instances of this model forming a prediction on images from the validation set are presented. We observe that, in all cases examined, the model follows a coherent reasoning process, even in cases where the model was confounded. Figure \ref{fig:local_17} shows a Least Auklet being correctly classified, Figure \ref{fig:local_253} illustrates a Horned Lark being correctly classified, and Figure \ref{fig:local_411} shows a Scarlet Tanager being correctly classified. 

\begin{figure}
    \centering
    \includegraphics[width=\textwidth]{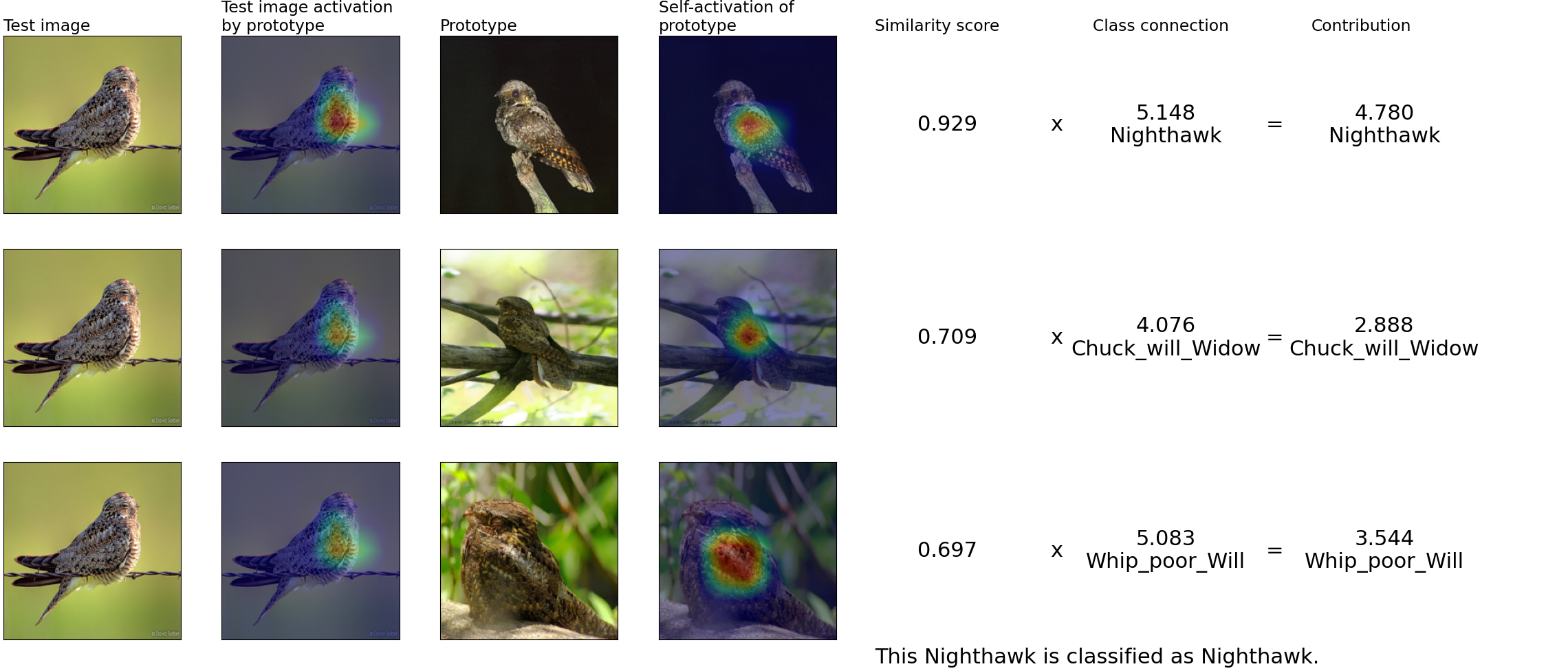}
    \caption{Reasoning process for the best model trained to jointly optimize accuracy and interpretability on an image of a Nighthawk. The model correctly classifies this image.}
    \label{fig:local_274}
\end{figure}

\begin{figure}
    \centering
    \includegraphics[width=\textwidth]{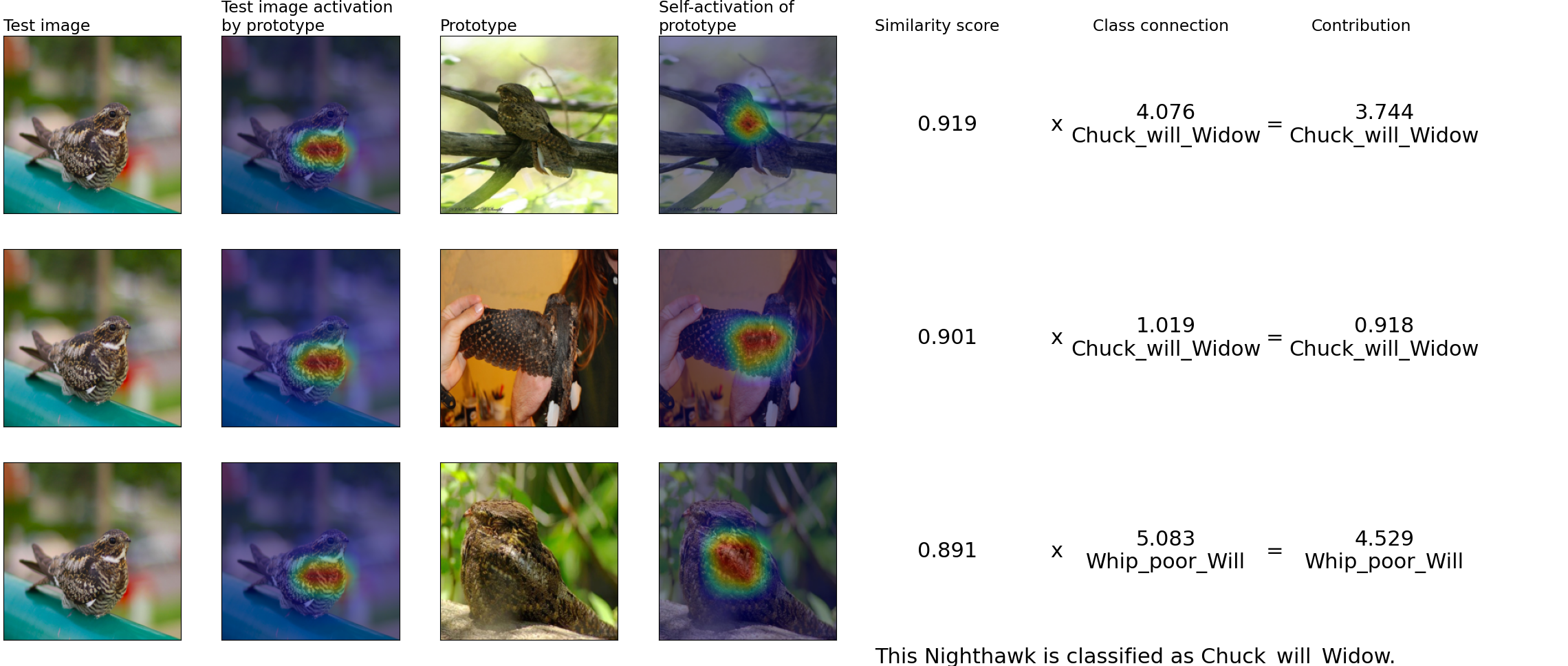}
    \caption{Reasoning process for the best model trained to jointly optimize accuracy and interpretability on an image of a Nighthawk. The model incorrectly classifies this image as a Chuck-will's-Widow, apparently confounded by the speckled pattern found on the wing of a Chuck-will's-Widow, which the model considers similar to the speckled breast of the Nighthawk.}
    \label{fig:local_275}
\end{figure}

Figure \ref{fig:local_274} illustrates a Nighthawk being correctly classified, while 
Figure \ref{fig:local_275} illustrates a Nighthawk being incorrectly classified as a Chuck-will's-Widow. In the first case, the model correctly identified the complicated speckled pattern on the Nighthawk's wing as being similar to a prototypical Nighthawk. In contrast, in Figure \ref{fig:local_275} the model confuses the speckled breast of a Nighthawk with the similarly speckled wing of a Chuck-will's-widow.

Several hundred additional visualizations from this model can be found here: \url{https://drive.google.com/drive/folders/13yQndNbLJiclv90UPwzemZ3rei2gzVty}.

%%%%%%%%%%%%%%%%%%%%%%%%%%%%%%%%%%%%%%%%%%%%%%%%%%%%%%%%%%%%

\clearpage
\newpage

\end{document}